# TA-RNN: an Attention-based Time-aware Recurrent Neural Network Architecture for Electronic Health Records

Mohammad Al Olaimat[1,3], Serdar Bozdag[1,2,3,*] and for the Alzheimer's Disease Neuroimaging Initiative [**]

[1] Dept. of Computer Science and Engineering, University of North Texas, Denton, USA, [2] Dept. of Mathematics, University of North Texas, Denton, USA, [3] BioDiscovery Institute, University of North Texas, Denton, USA. [**] Data used in preparation of this article were obtained from the Alzheimer's Disease Neuroimaging Initiative (ADNI) database (adni.loni.usc.edu). As such, the investigators within the ADNI contributed to the design and implementation of ADNI and/or provided data but did not participate in analysis or writing of this report. A complete listing of ADNI investigators can be found at: http://adni.loni.usc.edu/wp-content/uploads/how_to_apply/ADNI_Acknowledgement_List.pdf.

*To whom correspondence should be addressed.

## Abstract

**Motivation:** Electronic Health Records (EHR) represent a comprehensive resource of a patient's medical history. EHR are essential for utilizing advanced technologies such as deep learning (DL), enabling healthcare providers to analyze extensive data, extract valuable insights, and make precise and data-driven clinical decisions. DL methods such as Recurrent Neural Networks (RNN) have been utilized to analyze EHR to model disease progression and predict diagnosis. However, these methods do not address some inherent irregularities in EHR data such as irregular time intervals between clinical visits. Furthermore, most DL models are not interpretable. In this study, we propose two interpretable DL architectures based on RNN, namely Time-Aware RNN (TA-RNN) and TA-RNN-Autoencoder (TA-RNN-AE) to predict patient's clinical outcome in EHR at next visit and multiple visits ahead, respectively. To mitigate the impact of irregular time intervals, we propose incorporating time embedding of the elapsed times between visits. For interpretability, we propose employing a dual-level attention mechanism that operates between visits and features within each visit.
**Results:** The results of the experiments conducted on Alzheimer's Disease Neuroimaging Initiative (ADNI) and National Alzheimer's Coordinating Center (NACC) datasets indicated superior performance of proposed models for predicting Alzheimer's Disease (AD) compared to state-of-the-art and baseline approaches based on F2 and sensitivity. Additionally, TA-RNN showed superior performance on Medical Information Mart for Intensive Care (MIMIC-III) dataset for mortality prediction. In our ablation study, we observed enhanced predictive performance by incorporating time embedding and attention mechanisms. Finally, investigating attention weights helped identify influential visits and features in predictions.
**Availability:** https://github.com/bozdaglab/TA-RNN.
**Contact:** Serdar.Bozdag@unt.edu

## 1 Introduction

EHR represent a comprehensive resource that provide a historical collection of a patient's medical record and health-related data. This holistic resource includes structured data, such as patient medical conditions, medications, clinical measurements, and demographics, and unstructured data, exemplified by clinical notes (Hossain, Rana, Higgins, Soar, Barua, Pisani, D, et al., 2023; Lyu et al., 2022; Yadav et al., 2018). EHR have become the cornerstone in modeling classification of patients and progression and sub-typing of diseases through the advanced technologies (e.g., machine and deep learning), enabling healthcare providers to analyze vast volumes of data, extract valuable insights, and make more precise and data-driven clinical decisions (Xu et al., 2022; X. Yang et al., 2022).

Structured EHR are longitudinal in nature, providing a dynamic and chronological representation of a patient's medical history. The longitudinal nature of EHR reflects their capacity to capture and document a patient's health information over an extended time. Unlike static data snapshots, the temporal patterns embedded in EHR provide valuable insights into the progression of a patient's health. This, in turn, enables AI-based methods to make precise clinical decisions. However, machine learning methods such as Random Forest (RF) (Breiman, 2001), Support Vector Machine (SVM) (Cortes & Vapnik, 1995), and neural networks (McCulloch & Pitts, 1943) lack the capacity to handle temporal relations in EHR. These methods usually rely on a specific time point within the EHR, such as the baseline or the most recent clinical visit. Alternatively, decision can be made by aggregating data across all time points. Furthermore, EHR present data analysis challenges such as varying numbers of visits for patients, irregular time intervals between visits, and the presence of missing values. Consequently, AI-based methods for modeling EHR need to address these inherent issues.

To preserve the temporal nature of EHR and address varying number of visits per patient, recurrent neural networks (RNN) models such as Long Short-Term Memory (LSTM) (Hochreiter & Schmidhuber, 1997) and Gated Recurrent Unit (GRU) (Cho et al., 2014), along with Transformer (Vaswani et al., 2017) have been employed (Herp et al., 2023; Hossain, Rana, Higgins, Soar, Barua, Pisani, & Turner, 2023; Xu et al., 2022; Yadav et al., 2018; S. Yang et al., 2023; X. Yang et al., 2022).

In (G. Lee et al., 2019), a multimodal GRU-based approach was employed on EHR data related to Alzheimer's Disease (AD) to predict conversion from Mild Cognitive Impairment (MCI) to AD. In (Li & Fan, 2019), an LSTM-based deep learning model was proposed for early prediction of AD using EHR data related to AD. In (Nguyen, He, An, Alexander, Feng, & Yeo, 2020), an RNN-based model was applied to AD EHR data to predict the diagnosis of patients up to six years. In (Venugopalan et al., 2021), an integrative classification approach was proposed for early detection of AD stage. In (Fouladvand et al., 2021), a multi-stream Transformer-based approach was applied on EHR to predict opioid use disorder. Although RNN- and Transformer-based methods handle longitudinal EHR data, they do not consider irregular time intervals between clinical visits. RNN (e.g., LSTM and GRU) assume that temporal gaps between time points are equal while absolute position encoding is used in the original Transformer architecture (Vaswani et al., 2017).

Recent studies have proposed to extend LSTM to handle irregular time intervals. Time-aware LSTM (T-LSTM) was developed with the purpose of learning a subspace decomposition of the cell memory, enabling time decay to discount the memory content according to the elapsed time. (Baytas et al., 2017). In a subsequent study (Luong & Chandola, 2019), T-LSTM's effectiveness was assessed on a synthetic data and real EHR data from kidney patients. Despite its success in handling synthetic data, the results showed challenges in effectively sub-typing chronic kidney disease using the real EHR data. In (Liu et al., 2022), KIT-LSTM expands the LSTM model by incorporating two time-aware gates: one for the time gap between two consecutive visits and another for the time gap between consecutive measurements for each clinical feature. In (Al Olaimat et al., 2023), an RNN-based deep learning architecture called Predicting Progression of AD (PPAD) was proposed. PPAD addresses irregular time intervals through using patients' age in each visit as a feature to indicate time changes between successive visits. Although most recent RNN-based studies were able to mitigate the irregular time intervals, they exhibit limited interpretability, making the explanation of their prediction results challenging.

To improve model's interpretability while considering irregular time intervals, RETAIN (Choi et al., 2016) was proposed. This model aims to identify the significant time points and features influencing the prediction of heart disease using EHR. The methodology incorporates two attention mechanisms, facilitated by two distinct GRUs, to capture interactions between time points and features. Additionally, to address irregular time intervals, time information, such as duration between consecutive visits or cumulative number of days since the initial visit, are optionally introduced as an additional feature at each time point. Although the tool is available, its application is limited to MIMIC-III data, requiring algorithm reimplementation to use with different data formats. Furthermore, the effectiveness of incorporating time information as an additional feature remains unassessed. In (Zhang et al., 2019), another RNN-based architecture called ATTAIN was proposed to handle irregular time intervals. This approach incorporates both time intervals and an attention mechanism. The current prediction is formulated by aggregating information from all or some of the preceding memory cells, with regularization of the most recent memory cell guided by weights generated through the attention mechanism and time intervals, employing a time decay function. DATA-GRU (Tan et al., 2020) was proposed to predict patients' mortality risk. This model incorporates a time-aware mechanism to handle irregular time intervals through internally incorporating time intervals into DATA-GRU to adjust the hidden status in the previous memory cell. Additionally, a dual-attention mechanism is employed to address missing values by considering both data-quality and medical-knowledge perspectives. However, a notable limitation of this model lies in the excessive measures taken to address missing values and time intervals problem, leading to an increase in the model's complexity. This includes aspects such as the number of RNN cells employed, the number of learnable parameters, and the preprocessing steps conducted before training to generate the unreliability scores of the data. The Multi-Integration Attention Module (MIAM) (Y. Lee et al., 2022) was proposed for different downstream tasks to capture complex missing patterns in EHR. This involves combining missing indicators and time intervals, followed by integrating observations and missing patterns within the representation space using a self-attention mechanism.

Nevertheless, these approaches do not embed the time intervals directly into the visits data. Instead, they introduce them as additional features to handle irregular time intervals or internally modify the hidden states within RNN by updating the memory cells with the time intervals. Furthermore, most of these approaches lack a mechanism to interpret their results.

In this study, we present two interpretable DL architectures based on RNN, namely Attention-based Time-aware RNN (TA-RNN) and TA-RNN-Autoencoder (TA-RNN-AE) for early prediction of patient clinical outcomes at next visit and multiple visits ahead, respectively. To mitigate the effect of irregular time intervals, we propose incorporating elapsed times between consecutive visits into visit data through a time embedding layer. Additionally, to enhance the interpretability of the model's results, we propose utilizing a dual-level attention mechanism that operates between visits and features within each visit to identify the significant visits and features influencing the model's prediction. To demonstrate the robustness of our proposed models, we evaluated them on three real-world datasets: (i) early prediction of conversion to AD by training and testing the models on the Alzheimer's Disease Neuroimaging Initiative (ADNI) dataset; (ii) early prediction of conversion to AD by training the models on the ADNI dataset and testing them on the National Alzheimer's Coordinating Center (NACC) dataset; (iii) prediction of mortality by training and testing the models on the Medical Information Mart for Intensive Care (MIMIC-III) dataset. In the task of predicting AD conversion, our experiments demonstrated that our proposed models outperformed all baseline models across most prediction scenarios, particularly in terms of F2 and sensitivity. Regarding the mortality prediction task, the results indicated that TA-RNN outperformed RETAIN in AUC. We also illustrated that integrating time embedding into the input data contributed to improved model performance by effectively handling irregular time intervals between consecutive visits. Furthermore, the results demonstrated the remarkable ability of the dual-level attention mechanism to interpret the results of our proposed models.

The main contributions of this study can be summarized as follows:

- We propose two interpretable DL architectures based on RNN, namely Attention-based Time-aware RNN (TA-RNN) and TA-RNN-Autoencoder (TA-RNN-AE) for early prediction of patient clinical outcomes at the next visit and multiple visits ahead, respectively.

- We propose incorporating time embedding of the elapsed times between consecutive visits into visits' data through the utilization of a time embedding layer. This approach aims to mitigate the impact of the irregular time intervals.
- We propose employing a dual-level attention mechanism that operates between visits and features within each visit to identify notable visits and features influencing the model's predictions. This approach is intended to improve the interpretability of the model's results.
- We demonstrate that the proposed methods exhibit superior performance compared to the state-of-the-art (SOTA) and baseline approaches in downstream tasks. This was achieved by leveraging longitudinal multimodal data and cross-sectional demographic data from two large AD datasets, namely ADNI and NACC, for early prediction of conversion to AD. Additionally, we validated the effectiveness of our methods on a real-world EHR dataset obtained from MIMIC-III for predicting mortality.

## 2 Materials and methods

### 2.1 Datasets

In this study, three datasets were used to evaluate the proposed models. The first dataset consisted of longitudinal and cross-sectional data from the ADNI database (adni.loni.usc.edu). The ADNI was launched in 2003 as a public–private partnership, led by Principal Investigator Michael W. Weiner, MD. The primary goal of ADNI has been to test whether serial Magnetic Resonance Imaging (MRI), Positron Emission Tomography (PET), other biological markers, and clinical and neuropsychological assessment can be combined to measure the progression of MCI and early AD. Since it has been launched, the public–private cooperation has contributed to significant achievements in AD research by sharing data to researchers from all around the world (Jack et al., 2010; Jagust et al., 2010; Risacher et al., 2010; Saykin et al., 2010; Trojanowski et al., 2010; Weiner et al., 2010, 2013). The second dataset consisted of longitudinal and cross-experimental setup, we split the ADNI dataset into 70% training and 30% testing in a random stratified manner for model training and testing to predict conversion from MCI to AD, respectively. The ADNImerge R package (available at https://adni.bitbucket.io/) was employed to gather longitudinal and cross-sectional data from all ADNI studies, including ADNI1, ADNI2, and ADNI-GO (Jiang et al., 2020). We have conducted a preprocessing using the same steps as implemented in the PPAD (Al Olaimat et al., 2023). Briefly, in the preprocessing phase, we performed the removal of irrelevant features and visits, handled missing values through data imputation using K-Nearest Neighbors (KNN), and normalized the features. Following the preprocessing, the final dataset comprised 20 longitudinal and five cross-sectional demographic features for a total of 1169 patients and 5759 visits (Supplemental Table 1 and 2). For generalization, this entire process was iterated across three random splits.

In the second experimental setup, the entire ADNI dataset served as the training data, while the NACC dataset served as an external dataset for model testing to predict conversion from MCI to AD. We conducted a data harmonization step to collect common features from the ADNI and the NACC datasets. Subsequently, the harmonized datasets were preprocessed, following the same steps as outlined in the PPAD (Al Olaimat et al., 2023). Ultimately, we successfully harmonized nine features between the ADNI and NACC datasets. The final ADNI dataset consisted of 1205 patients and 6066 visits (Supplemental Table 3 and 4). The difference in patient and visit numbers between the first and second experimental setups in the ADNI data arises due to the preprocessing step. Since the second experimental setup used fewer features, the missing rate was lower, hence more visits were kept. The final NACC dataset encompassed 8121 patients and 35,423 visits.

For the third experimental setup, the MIMIC-III dataset was utilized to evaluate the proposed models for mortality prediction task. We extracted patients' visits, mortality labels, and time information from the MIMIC-III dataset using the same procedures that have been employed in RETAIN (Choi et al., 2016). Patients' visits include the International Classification of Diseases, Ninth Revision (ICD-9) codes, with dataset includes 942 unique ICD-9 codes for 7537 patients. To train the proposed models, the sectional data from the NACC database (Besser et al., 2018). The NACC database is a centralized asset for AD research, designed specifically to facilitate and expedite investigations into the causes, diagnosis, and treatment of AD. It comprises information from various study sites throughout the United States, encompassing a range of data such as demographics, cognitive assessments, genetic details, and MRI data. The third dataset was from the MIMIC-III database (Goldberger et al., 2000; A. Johnson et al., 2016; A. E. W. Johnson et al., 2016). The MIMIC-III database serves as a comprehensive resource of EHR designed for researchers interested in gaining insights into critical care practices and patient clinical outcomes, with a specific focus on patients of the Intensive Care Unit (ICU). The MIMIC-III databases consist of multiple set of clinical data, including vital signs, laboratory results, conditions, medical procedures, medications, and clinical notes.

*Table 1. Experimental setups that were employed for the proposed models. In the first and second experimental setups, 5-fold cross-validation was employed on the training data to tune the hyperparameters.*

| Experimental setup | Training data | Validation data | Test data | Downstream task |
|---|---|---|---|---|
| First | 70% of ADNI | 5-fold cross-validation | 30% of ADNI | Predicting conversion of MCI to AD |
| Second | Entire ADNI | 5-fold cross-validation | Entire NACC | Predicting conversion of MCI to AD |
| Third | 70% of MIMIC-III | 10% of MIMIC-III | 20% of MIMIC-III | Predicting mortality |

We employed three distinct experimental setups to train and evaluate our proposed models utilizing these three datasets (Table 1). In the first dataset underwent a stratified random split, allocating 70% for training, 20% for testing, and 10% for validation (Supplemental Table 9).

In all experimental setups, datasets have been preprocessed such that each sample represents a unique patient with at least two visits. Consequently, the input data is structured in three dimensions, delineated as (samples, visits, and features).

### 2.1.1 Dataset notations

Let $D$ denote the longitudinal EHR data with $N$ samples (i.e., patients). $D = (X_1, X_2, ..., X_N)$ where each sample $X$ represents measurements of $F$ features collected over $T$ time points (i.e., visits): $X = \{x_1, x_2, ..., x_T\} \in \mathbb{R}^{T \times F}$. For each visit $t \in \{1, 2, ..., T\}$, $x_t = \{x_t^1, x_t^2, ..., x_t^F\} \in \mathbb{R}^F$ represents a vector of features of sample $X$ at visit $t$. For each feature $f \in \{1, 2, ..., F\}$, $x^f = \{x_1^f, x_2^f, ..., x_T^f\} \in \mathbb{R}^T$ represents the $f$th feature value of sample $X$ over $T$ visits, and $x_t^f$ represents the $f$th feature value of sample $X$ at visit $t$. In $D$, each sample has a corresponding elapsed time data $(E_1, E_2, ..., E_N)$, where each $E$ represents the elapsed times for sample $X$ collected over $T$ time points (visits): $E = \{e_1, e_2, ..., e_T\} \in \mathbb{R}^{T \times 1}$. For each visit $t \in \{1, 2, ..., T\}$, $e_t \in \mathbb{R}$ represents the elapsed time at visit $t$, such that $e_t = \begin{cases} 0, & \text{if } t = 1 \\ vd_t - vd_{t-1}, & \text{if } t > 1 \end{cases}$, where $vd_t$ represents the visit date at visit $t$. For ADNI and NACC data, the unit of $e$ was in years, whereas for MIMIC-III data, the unit was in days. Finally, in $D$ each sample has a corresponding diagnosis or clinical outcome $(Y_1, Y_2, ..., Y_N)$ for each time point: $Y = \{y_1, y_2, ..., y_T\} \in \mathbb{R}^{T \times 1}$. In this study, for each visit $t \in \{1, 2, ..., T\}$, $y_t \in \{0, 1\}$ where 0 denotes MCI and 1 denotes AD in the MCI to AD conversion prediction task, and 0 denotes absence of mortality and 1 denotes mortality in the mortality prediction task.

## 2.2 The proposed method

In this study, we developed two interpretable RNN-based DL architectures to predict the clinical outcome at the next visit and the multiple visits ahead. To demonstrate the robustness of the proposed models, we evaluated them using two downstream tasks, namely predicting conversion from MCI to AD and patient mortality prediction. In both architectures, we employed a time embedding layer that incorporates elapsed time between consecutive visits in EHR into input data to address lack of consideration of irregular time intervals between consecutive inputs by RNN models. Furthermore, we employed a dual-level attention mechanism that operates between visits and features within each visit to identify significant visits and features influencing predictions. Implementing the dual-level attention mechanism, in turn, improves the interpretability of the model. Finally, the RNN cell type was considered as a hyperparameter, offering options such as LSTM, GRU, bidirectional LSTM (Bi-LSTM), and bidirectional GRU (Bi-GRU).

### 2.2.1 Time embedding layer

To alleviate the limitation of RNN models with irregular time intervals, we employed a time embedding layer. The layer takes the longitudinal ($X$) and elapsed time ($E$) data as inputs to generate a new representation in a manner where the time information is integrated with the original data. This layer is a variation of positional encoding for continuous time values as input (Y. Lee et al., 2022). It transforms these values into an encoded vector representation ($TE$) (Eq 1), which is an adapted version of the positional encoding formula proposed in (Vaswani et al., 2017).

$$TE_{(e,i)} = \begin{cases} sin\left(\dfrac{e}{ET_{max}^{2i/d_{model}}}\right) & \text{if } i \text{ is even} \\ cos\left(\dfrac{e}{ET_{max}^{2i/d_{model}}}\right) & \text{if } i \text{ is odd} \end{cases} \qquad (1)$$

In Eq 1, $e$ represents the elapsed time since the previous visit, $i$ represents the index in the embedding space, $d_{model}$ represents the model dimension or size, and $ET_{max}$ represents the maximum elapsed of time data. Then, the longitudinal input data $X$ is added to the time embedding $TE$ to generate a new embedding or representation ($Z$) (Eq 2) of $X$ that has time information based on $E$.

$$Z = X + TE \qquad (2)$$

In Eq 2, $Z = \{z_1, z_2, ..., z_T\} \in \mathbb{R}^{T \times d_{model}}$ represents the new embedding for the input data that incorporates the temporal representation. In the following sections $Z$ will be the input to the proposed models.

### 2.2.2 A predictive model for clinical outcome at next visit

TA-RNN comprises three components, namely, a time embedding layer, attention-based RNN, and a prediction layer based on multi-layer perceptron (MLP) (Figure 1). In this model, to address the irregular time intervals, first, the time embedding layer is used to generate the new input embedding $Z$ based on the original input data $X$ and time embedding (Eq 2,

Section 2.2.1). Then, $Z$ is fed to an RNN cell (Eq 3) that employs a dual sets of attention weights (for visits and features) and generates the hidden state $h_t \in \mathbb{R}^{hidden\_size}$ for each time point $t$.

$$h_1, .., h_t = RNN(z_1, ..., z_t) \qquad (3)$$

To control the impact of each visit's embedding $(z_1, ..., z_t)$, the scalars $(\alpha_1, ..., \alpha_t)$ that represent visit-level attention weights are calculated using hidden states of the RNN cell (Eq 4).

$$k_j = W_\alpha h_j + b_\alpha \quad for \ j = 1, ..., t$$
$$\alpha_1, ..., \alpha_t = softmax(k_1, ..., k_t) \qquad (4)$$

In Eq 4, $W_\alpha \in \mathbb{R}^{hidden\_size}$ and $b_\alpha \in \mathbb{R}$ are the trainable parameters. On the other hand, to control the impact of features in each visit's embedding $(z_1, ..., z_t)$, the vectors $(\beta_1, ..., \beta_t)$ that represent feature-level attention weights are calculated using hidden states of the RNN cell (Eq 5) where $W_\beta \in \mathbb{R}^{hidden\_size \times d_{model}}$ and $b_\beta \in \mathbb{R}^{d_{model}}$ are the trainable linear transformation matrix and the bias vector, respectively.

$$\beta_j = tanh(W_\beta h_j + b_\beta) \ for \ j = 1, ..., t$$
$$\beta_j = softmax(\beta_j^1, \beta_j^2, ..., \beta_j^{d_{model}}) \, for \, j = 1, .., t \qquad (5)$$

Then, the obtained visits and features attention weights are utilized to generate the context vector ($c_t \in \mathbb{R}^{d_{model}}$) (Eq 6) of a sample in the EHR up to the $t^{th}$ visit.

$$c_t = \sum_{j=1}^{t} \alpha_j \, \beta_j \odot z_j \qquad (6)$$

In Eq 6, $\odot$ represents the element-wise multiplication operation. Finally, if the demographic data ($Dem$) is available, the $c_t$ is concatenated with it to train an MLP for predicting the clinical outcome of the next visit (Eq 7).

$$y' = \sigma\big(W_1(ReLU(W_2(c \oplus Dem) + b_2)) + b_1\big) \qquad (7)$$

In Eq 7, $y'$ represents the predicted clinical outcome, $W_1$ and $W_2$ are the trainable linear transformation matrices, $b_1$ and $b_2$ are the trainable bias vectors, $\sigma$ represents the sigmoid function, and $\oplus$ represents the concatenation operation.

### 2.2.3 A predictive model for clinical outcome at multiple visits ahead

To predict clinical outcome in EHR at multiple visits ahead, we developed another predictive model called Time-Aware RNN Autoencoder (TA-RNN-AE). TA-RNN-AE comprises time embedding, attention-based RNN autoencoder, and MLP (Figure 2). In this model, the encoder component learns $c_t$ of the longitudinal data as described in the previous section (Eq 6). Subsequently, the decoder component generates multiple upcoming visits' representations using $c_t$ up to $n$ visits in autoregressive manner. Initially, both the input and the initial hidden state to the decoder are set to $c_t$. Subsequently, for each subsequent time step, the input to the decoder is derived from the output of the previous time step (Eq 8).

$$g_1, g_2 ..., g_n = RNN(c, g_1, ..., g_{n-1}) \qquad (8)$$

Finally, if demographic data $Dem$ is available, the MLP model is then trained by concatenating it with the last visit's representation generated by the decoder $x_{t+n}$. This training is designed to predict the clinical outcome in the EHR at the $(t + n)^{th}$ visit, as outlined in (Eq 9).

$$y' = \sigma(W_1(ReLU(W_2(x_{t+n} \oplus Dem) + b_2)) + b_1) \quad (9)$$

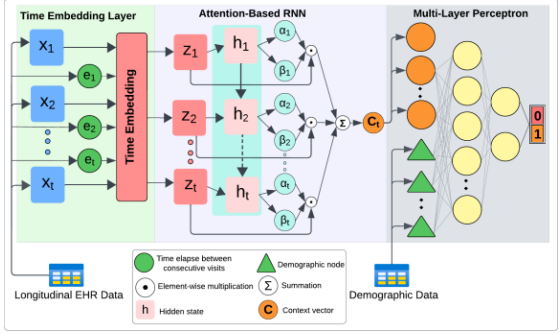

*Figure 1.* TA-RNN architecture for predicting clinical outcomes in EHR at next visit.

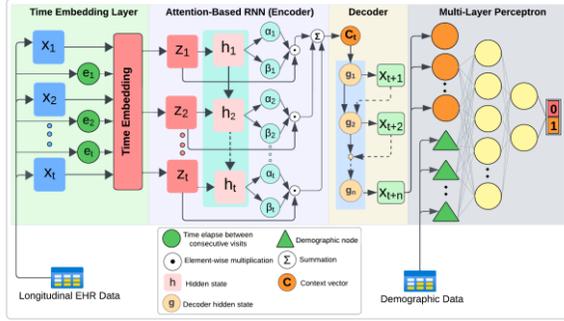

*Figure 2.* TA-RNN-AE architecture for predicting of clinical outcome in EHR at multiple visits ahead.

### 2.2.4 Parameter learning and evaluation metrics

In both proposed architectures, trainable parameters of all components are jointly learned using a customized binary cross-entropy loss function (Eq 10). This customized loss function assigns greater weight to predicting positive cases to minimize predicting false negative cases which in turn increases the sensitivity of the predictive model.

$$Loss = -\frac{1}{N}\sum \left( \delta \cdot (y \cdot log\, y') \right) + \left( (1 - \delta) \cdot (1 - y) \cdot log(1 - y') \right) \quad (10)$$

In Eq 9, $\delta$ is a hyperparameter real number between 0 and 1 to define the relative weight of positive prediction and $y$ is the true diagnosis. In this study, we set $\delta$ to 0.7 and 0.65 for predicting AD conversion and mortality tasks, respectively. For the optimization process, all models underwent training with the Adaptive Moment Estimation (Adam) optimizer (Kingma & Ba, 2014), utilizing the default learning rate 0.001. Hyperparameters such as the RNN cell type, number of epochs, batch size, dropout rate, L2 regularization, hidden size, embedding size, and $\delta$ were tuned using 5-fold cross validation and 10% validation data for predicting AD conversion and mortality tasks, respectively. Model evaluation was performed using the F2 score (Eq 11) and sensitivity.

$$F_\beta = (1 + \beta^2) \cdot \frac{precision \cdot recall}{\beta^2 . precision + recall} \quad (11)$$

In Eq 11, recall is given $\beta$ times more emphasis than precision. We set $\beta$ to 2 in this study.

## 3    Results and discussion

This study proposes two RNN-based frameworks: TA-RNN and TA-RNN-AE to predict the clinical outcome in EHR at next visit and at multiple visits ahead, respectively. The proposed frameworks were evaluated based on three experimental setups (see Section 2.1 for details). We also conducted an ablation study to assess the importance of the time embedding and dual-attention components. Finally, we have assessed the interpretability of the model by examining the visit- and feature-based attention weights for the entire test set and for an individual patient. The details of these analyses are described below.

### 3.1 Predicting clinical outcome at next visits

To evaluate the predictive performance of TA-RNN, in the first and second experimental setup, we trained four models using two, three, five, and six preceding visits to predict the conversion to AD at the subsequent visit. For hyperparameter tuning, grid search with 5-fold cross-validation was performed. The optimal hyperparameters of the models in the first and second experimental setup are shown in Supplemental Table 5 and 6, respectively.

We compared TA-RNN to RF, SVM, T-LSTM, and PPAD. PPAD was trained using the longitudinal features including the age while RF and SVM, which are not designed for longitudinal data, were trained on the aggregated longitudinal data after averaging each feature's values across all visits. On the other hand, T-LSTM and TA-RNN were trained without utilizing the age feature to evaluate the effectiveness of their mechanisms of handling irregular time intervals. The entire process was repeated 15 times in the second experimental setup and for each of the three random data splits in the first experimental setup to ensure generalizability. The outcomes in the first and second experimental setups demonstrated the superiority of TA-RNN upon all baseline models in terms of F2 (Fig. 3A and 3B) and sensitivity (Supplemental Fig. 1A and 1B) except one case in the second experimental setup (Fig. 3B) and (Supplemental Fig. 1B). Moreover, this superior performance highlights TA-RNN's capability to address the irregular time intervals issue better than T-LSTM and PPAD. We observed that the performance difference between TA-RNN and RF, SVM, and T-LSTM were statistically significant (t-test $p$-value $\leq 0.05$) for all cases except for one. Although the performance difference between TA-RNN and PPAD was not statistically significant in most cases (Supplemental Table 11 and 12), TA-RNN's interpretability feature through its dual-level attention mechanism is a clear advantage over PPAD.

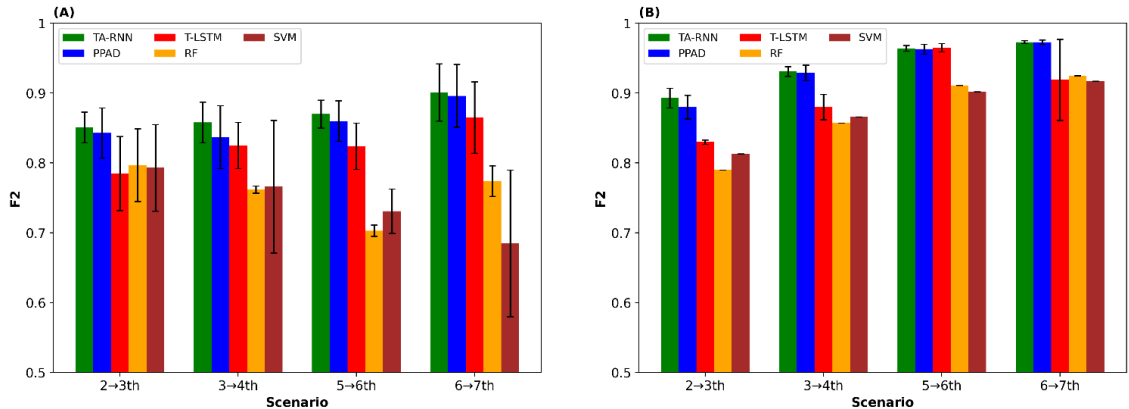

***Figure 3.*** *F2 scores for TA-RNN models predicting conversion to AD at the next visit. (A) Models evaluated on held-out samples in ADNI after training with two, three, five, and six preceding visits in ADNI. (B) Models evaluated on NACC after training with two, three, five, and six preceding visits in ADNI. Error bars represent standard deviations. Statistical significance between TA-RNN and other tools was assessed using t-test (see Supplemental Table 11 and 12 for p-values).*

For the third experimental setup, our evaluation focused only on TA-RNN and excluded TA-RNN-AE as 70% and 18% of patients in the MIMIC-III dataset have only two and three visits, respectively. TA-RNN was trained on MIMIC-III data and compared only with RETAIN, which was designed to enhance model's interpretation, for mortality prediction at the last hospital stay. For hyperparameter tuning, 10% validation data was utilized. The optimal hyperparameters are shown in Supplemental Table 10. The entire process was repeated 15 times to ensure generalizability. We observed that TA-RNN outperformed the RETAIN model in AUC slightly (Table 2). Although the difference between their performances was marginally significant ($p$-value = 0.05), TA-RNN exhibits lower complexity compared to RETAIN as TA-RNN utilizes only one RNN cell. On the other hand, RETAIN employs two separate RNN cells – one for handling visit attention weights and the other for handling feature attention weights. Reduced complexity makes TA-RNN faster to train with less trained data compared to RETAIN.

***Table 2***. *AUC score for TA-RNN model for mortality prediction at the next visit.*

| Model | AUC |
|---|---|
| **RETAIN** | $0.731 \pm 0.002$ |
| **TA-RNN** | $\mathbf{0.733 \pm 0.003}$ |

### 3.2 Predicting clinical outcome at multiple visits ahead

To evaluate the predictive performance of TA-RNN-AE, we conducted the training for various scenarios, including models trained on two, three, five, and six visits to predict the conversion to AD at the subsequent second, third, and fourth visits. This was accomplished using both the first and second experimental setup. For instance, a trained model on datasets from two clinical visits was evaluated for its ability to predict the conversion to AD at the fourth, fifth, and sixth visits. For hyperparameter tuning, grid search with 5-fold cross-validation was performed. The optimal hyperparameters of the models in the first and second experimental setup are shown in Supplemental 7 and 8, respectively. We also repeated the entire process 15 times in the second experimental setup and for each of the three random data splits in the first experimental setup to ensure generalizability. Because T-LSTM and RETAIN do not have the ability to predict the clinical outcome in EHR at multiple visits ahead, TA-RNN-AE comparison was limited to RF, SVM, and PPAD-AE. The results demonstrated the superiority of TA-RNN-AE upon all baseline and SOTA models in terms of F2 (Fig. 4) and sensitivity (Supplemental Fig. 2) except for one case where PPAD-AE outperformed TA-RNN-AE slightly in terms of F2 (Fig. 4D). The performance difference between TA-RNN-AE and RF and SVM was statistically significant (t-test $p$-value $\leq 0.05$) for 45 out of 48 cases in the first and second experimental setup. TA-RNN-AE outperformed PPAD-AE significantly for three and seven out of twelve scenarios in the first and second experimental setup, respectively (Supplemental Table 13 and 14). Furthermore, unlike PPAD-AE, TA-RNN-AE provides higher interpretability through its dual-level attention mechanism.

### 3.3 Ablation study

We conducted an ablation study for TA-RNN and TA-RNN-AE based on the first and second experimental setups to investigate the contribution of the time embedding layer that addresses the irregular time interval issue and the dual-level attention mechanism that improves the proposed models' interpretability. Here, we evaluated the predictive performance of the proposed models with two variations (i) without using the time embedding nor elapsed time, but using dual-level attention (i.e., A-RNN and A-RNN-AE); (ii) without using the dual-level attention mechanism but using time embedding (i.e., T-RNN and T-RNN-AE). We observed that the performance of A-RNN was lower compared to TA-RNN in all cases for first (Table 3) and the second experimental setup (Supplemental Table 15). The results also showed that the performance of T-RNN was slightly lower compared to TA-RNN in two out of four cases for the first experimental setup (Table 3) and three out of four cases for the second experimental setup (Supplemental Table 15). In addition, we observed that A-RNN-AE performed lower than TA-RNN-AE in seven out of twelve cases and remained unchanged in one case for the first experimental setup (Supplemental Table 16), and eleven out of twelve cases for the second experimental setup (Supplemental Table 17). Finally, the performance of T-RNN-AE was lower compared to TA-RNN-AE in all cases except for one case for the first experimental setup (Supplemental Table 16) and all cases for the second experimental setup (Supplemental Table 17). The ablation study showed a decrease in the performance of the proposed models across most scenarios, indicating the significance of the time embedding layer and dual-level attention mechanism in enhancing both the performance and interpretability of the models.

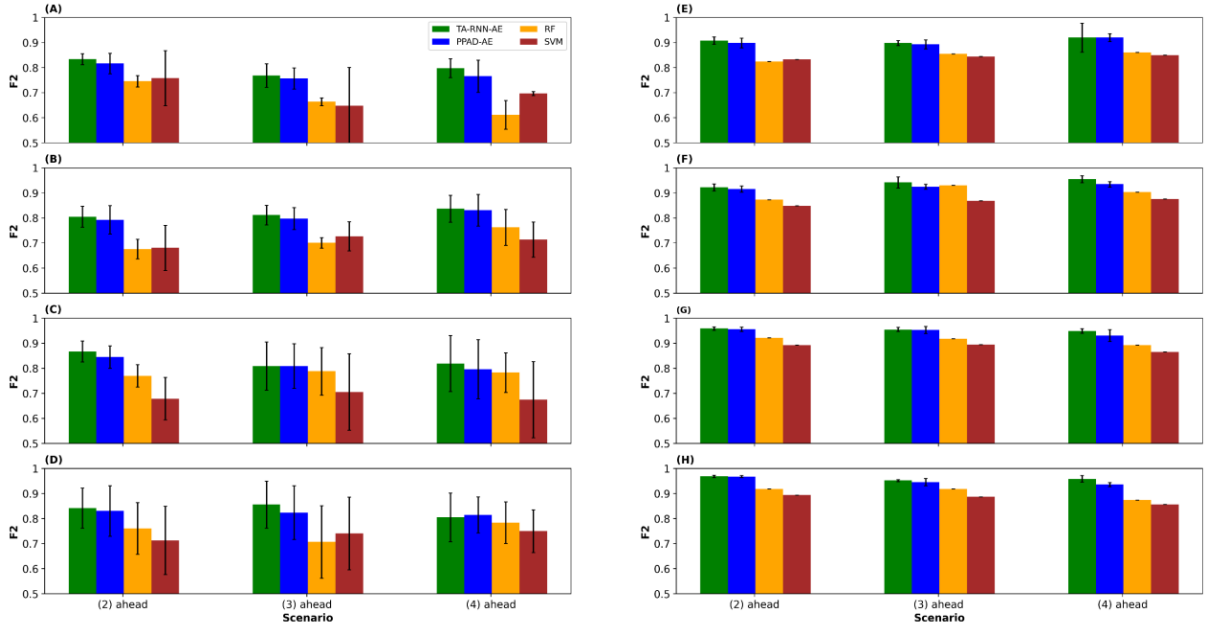

*Figure 4.* *F2 scores for TA-RNN-AE models predicting conversion to AD at the next second, third, and fourth visits ahead. (A, B, C, and D) Models were tested on held-out samples in ADNI after training using 2, 3, 5, and 6 visits in ADNI, respectively. (E, F, G, and H) Models were tested on NACC after training using 2, 3, 5, and 6 visits in ADNI, respectively. Error bars represent standard deviations. Statistical significance between TA-RNN-AE and other tools was assessed using t-test (see Supplemental Table 13 and 14 for p-values).*

*Table 3.* *TA-RNN F2 scores for the ablation of dual-level attention and time embedding. A-RNN: TA-RNN without time embedding. T-RNN: TA-RNN without dual-attention. TA-RNN was evaluated on held-out samples in ADNI after training using two, three, five, and six preceding visits in ADNI. Best F2 score in each case (column) is shown in bold.*

| Model \ Scenario | 2→1 | 3→1 | 5→1 | 6→1 |
|---|---|---|---|---|
| **A-RNN** | $0.844 \pm 0.040$ | $0.850 \pm 0.035$ | $0.855 \pm 0.029$ | $0.887 \pm 0.035$ |
| **T-RNN** | $\mathbf{0.852 \pm 0.031}$ | $0.855 \pm 0.019$ | $0.869 \pm 0.020$ | $\mathbf{0.905 \pm 0.036}$ |
| **TA-RNN (Proposed)** | $0.851 \pm 0.022$ | $\mathbf{0.858 \pm 0.029}$ | $\mathbf{0.870 \pm 0.020}$ | $0.901 \pm 0.041$ |

### 3.4 Interpretations of the models' predictions

To demonstrate the interpretability of the models, we visualized their behavior during the prediction process using the generated attention weights for visits and features. In the first experimental setup, for example, TA-RNN model that was trained on six visits to predict the clinical outcome in the next visit generated six visit-level attention weights (i.e., one for each visit). The attention weight values for each visit range from 0 to 1, and the sum of all six weights equals to 1. Additionally, the model generated 19 attention weights, one for each feature (except for the age feature, which was not utilized in our models) at each visit, representing the feature-level attention. These weights provide insights into the influence of each visit and feature during the model's prediction of the clinical outcome. Fig. 5, visualizing the attention weights at the visit level, demonstrates that for most samples the model prioritizes on the final visits—specifically the fifth and sixth visits—during predictions. Additionally, Fig. 6 shows the average attention weight at the feature level across all visits for the same example. We observed that CDRSB, MMSE, RAVLT.learning, and FAQ features exhibit the most substantial influence on the model's predictions, aligning with observations in the literature (Al Olaimat et al., 2023; Nguyen, He, An, Alexander, Feng, Yeo, et al., 2020; Perera & Ganegoda, 2023; Velazquez & Lee, 2021; Zhang et al., 2024). In addition, we evaluated TA-RNN's interpretability in predicting MCI to

AD conversion using a patient from the test dataset who converted from MCI to AD in the subsequent visit (7th), where the model accurately predicted the conversion. We made the same observation that fifth and sixth visits are the most influential visit for this patient (Fig. 7A).

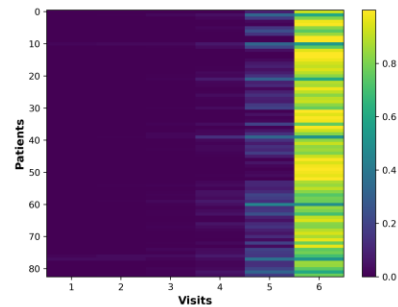

*Figure 5.* *Attention weights at the visit level for TA-RNN. TA-RNN was evaluated on held-out samples in ADNI after training using six preceding visits in ADNI.*

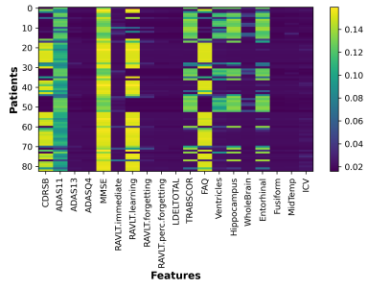

**Figure 6.** Average attention weights at the feature level for TA-RNN. TA-RNN was evaluated on held-out samples in ADNI after training using six preceding visits in ADNI.

Furthermore, several cognitive test features were important regardless all visits (Fig. 7B). After combining feature and visit attention values, the model primarily focuses on CSRSB, MMSE, RAVLT.learning, and FAQ features in the model's prediction (Fig. 7C). In light of the preceding observations, the dual-attention mechanism of our model holds significant promise for enhancing the interpretability and utility of our model.

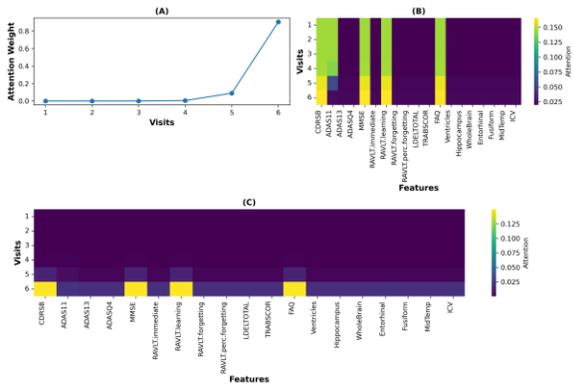

**Figure 7.** Visualization of TA-RNN's behavior during correctly predicting the conversion from MCI to AD for a patient. A, B, and C represents visits attention weights, features attention weights, and element-wise multiplication of attention weights, respectively. TA-RNN was evaluated on held-out samples in ADNI after training using six preceding visits in ADNI.

## 4   Conclusion

In this study, we propose two RNN-based architectures, namely TA-RNN and TA-RNN-AE for prediction of clinical outcome in EHR at next visit and multiple visits ahead, respectively. The proposed models were able to address the irregular time interval issue in longitudinal EHR through a time embedding layer that incorporates time intervals with original data. Additionally, model predictions were made interpretable by utilizing a dual-level attention mechanism that identifies the significant visits and features influencing each prediction. To evaluate the effectiveness of our proposed models, we employed three experimental setups using two extensive EHR datasets obtained from ADNI and NACC for predicting MCI to AD conversion, as well as data from MIMIC-III for mortality prediction. In all experimental setups, the results showed that TA-RNN and TA-RNN-AE outperformed baseline and SOTA methods in almost all cases. TA-RNN and TA-RNN-AE source code and documentation are publicly available at https://github.com/bozdaglab/TA-RNN/.

## Acknowledgments

Data collection and sharing for this project was funded by the Alzheimer's Disease Neuroimaging Initiative (ADNI) (National Institutes of Health Grant U01 AG024904) and DOD ADNI (Department of Defense award number W81XWH-12-2-0012). ADNI is funded by the National Institute on Aging, the National Institute of Biomedical Imaging and Bioengineering, and through generous contributions from the following: AbbVie, Alzheimer's Association; Alzheimer's Drug Discovery Foundation; Araclon Biotech; BioClinica, Inc.; Biogen; Bristol-Myers Squibb Company; CereSpir, Inc.; Cogstate; Eisai Inc.; Elan Pharmaceuticals, Inc.; Eli Lilly and Company; EuroImmun; F. Hoffmann-La Roche Ltd and its affiliated company Genentech, Inc.; Fujirebio; GE Healthcare; IXICO Ltd.; Janssen Alzheimer Immunotherapy Research & Development, LLC.; Johnson & Johnson Pharmaceutical Research & Development LLC.; Lumosity; Lundbeck; Merck & Co., Inc.; Meso Scale Diagnostics, LLC.; NeuroRx Research; Neurotrack Technologies; Novartis Pharmaceuticals Corporation; Pfizer Inc.; Piramal Imaging; Servier; Takeda Pharmaceutical Company; and Transition Therapeutics. The Canadian Institutes of Health Research is providing funds to support ADNI clinical sites in Canada. Private sector contributions are facilitated by the Foundation for the National Institutes of Health (www.fnih.org). The grantee organization is the Northern California Institute for Research and Education, and the study is coordinated by the Alzheimer's Therapeutic Research Institute at the University of Southern California. ADNI data are disseminated by the Laboratory for Neuro Imaging at the University of Southern California. The NACC database is funded by NIA/NIH Grant U24 AG072122. NACC data are contributed by the NIA-funded ADRCs: P30 AG062429 (PI James Brewer, MD, PhD), P30 AG066468 (PI Oscar Lopez, MD), P30 AG062421 (PI Bradley Hyman, MD, PhD), P30 AG066509 (PI Thomas Grabowski, MD), P30 AG066514 (PI Mary Sano, PhD), P30 AG066530 (PI Helena Chui, MD), P30 AG066507 (PI Marilyn Albert, PhD), P30 AG066444 (PI John Morris, MD), P30 AG066518 (PI Jeffrey Kaye, MD), P30 AG066512 (PI Thomas Wisniewski, MD), P30 AG066462 (PI Scott Small, MD), P30 AG072979 (PI David Wolk, MD), P30 AG072972 (PI Charles DeCarli, MD), P30 AG072976 (PI Andrew Saykin, PsyD), P30 AG072975 (PI David Bennett, MD), P30 AG072978 (PI Neil Kowall, MD), P30 AG072977 (PI Robert Vassar, PhD), P30 AG066519 (PI Frank aferla, PhD), P30 AG062677 (PI Ronald Petersen, MD, PhD), P30 AG079280 (PI Eric Reiman, MD), P30 AG062422 (PI Gil Rabinovici, MD), P30 AG066511 (PI Allan Levey, MD, PhD), P30 AG072946 (PI Linda Van Eldik, PhD), P30 AG062715 (PI Sanjay Asthana, MD, FRCP), P30 AG072973 (PI Russell Swerdlow, MD), P30 AG066506 (PI Todd Golde, MD, PhD), P30 AG066508 (PI Stephen Strittmatter, MD, PhD), P30 AG066515 (PI Victor Henderson, MD, MS), P30 AG072947 (PI Suzanne Craft, PhD), P30 AG072931 (PI Henry Paulson, MD, PhD), P30 AG066546 (PI Sudha Seshadri, MD), P20 AG068024 (PI Erik Roberson, MD, PhD), P20 AG068053 (PI Justin Miller, PhD), P20 AG068077 (PI Gary Rosenberg, MD), P20 AG068082 (PI Angela Jefferson, PhD), P30 AG072958 (PI Heather Whitson, MD), P30 AG072959 (PI James Leverenz, MD).

## Funding

This work was supported by the National Institute of General Medical Sciences of the National Institutes of Health under Award Number R35GM133657 and the startup funds from the University of North Texas.